# ERIENet: An Efficient RAW Image Enhancement Network under Low-Light Environment

1st Jianan Wang
School of Computer Science
Beijing Institute of Technology
Beijing, China
wangjianan01@bit.edu.cn

2nd Yang Hong
School of Computer Science
Beijing Institute of Technology
Beijing, China
hong66854380@gmail.com

3rd Hesong Li
School of Computer Science
Beijing Institute of Technology
Beijing, China
lihesong2@bit.edu.cn

4th Tao Wang
Department of Planning
Ministry of Emergency Management
Big Data Center
Beijing, China
784090912@qq.com

5th Songrong Liu
Zhejiang Communications Involvement
Expressway Operation Management
Co., Ltd.
Zhejiang, China
59036251@qq.com

6th Ying Fu*
School of Computer Science
Beijing Institute of Technology
Beijing, China
fuying@bit.edu.cn

*Abstract*—RAW images have shown superior performance than sRGB images in many image processing tasks, especially for low-light image enhancement. However, most existing methods for RAW-based low-light enhancement usually sequentially process multi-scale information, which makes it difficult to achieve lightweight models and high processing speeds. Besides, they usually ignore the green channel superiority of RAW images and fail to achieve better reconstruction performance with good use of green channel information. In this work, we propose an efficient RAW Image Enhancement Network (ERIENet), which parallelly processes multi-scale information with efficient convolution modules, and takes advantage of rich information in green channels to guide the reconstruction of images. Firstly, we introduce an efficient multi-scale fully-parallel architecture with a novel channel-aware residual dense block to extract feature maps, which reduces computational costs and achieves real-time processing speed. Secondly, we introduce a green channel guidance branch to exploit the rich information within the green channels of the input RAW image. It increases the quality of reconstruction results with few parameters and computations. Experiments on commonly used low-light image enhancement datasets show that ERIENet outperforms state-of-the-art methods in enhancing low-light RAW images with higher efficiency. It also achieves an optimal speed of over 146 frame-per-second (FPS) for 4K-resolution images on a single NVIDIA GeForce RTX 3090 with 24G memory.

## I. Introduction

Image enhancement and understanding technologies provide support for multiple fields [1], [2], [3], [4], [5], [6], [7], [8]. Among existing image enhancement methods, low-light image enhancement is crucial for solving low illumination and contrast problems. While deep neural networks have advanced this field, they often introduce noise or artifacts in extremely dark areas. To address this, methods now jointly perform brightness correction and noise removal [9]. Compared to sRGB images, RAW data retains more of the original sensor data, providing a higher dynamic range and more color detail, making it superior for reconstruction tasks. However, RAW image enhancement suffers from low efficiency due to high resolution and model complexity. Lightweight networks improve efficiency but often at the cost of quality [14]. Efficient, high-quality methods remain a challenge. Besides, in Bayer images, the sampling rate of green pixels is twice that of red and blue pixels, recording

This work was supported by the National Natural Science Foundation of China (62331006 and 62171038), and the Fundamental Research Funds for the Central Universities.

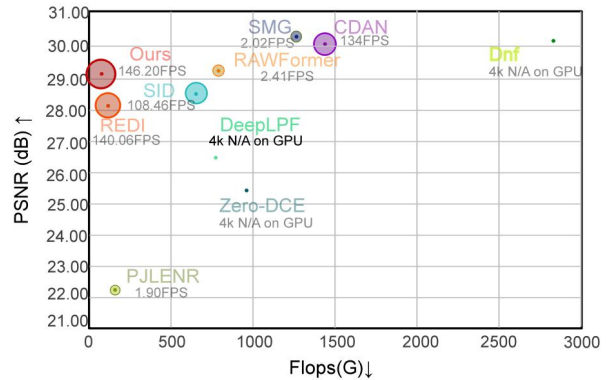

Fig. 1. PSNR, FLOPs, and Frame-Per-Second (FPS) of different methods on SID [9] dataset. The radius of each circle indicates the processing speed (FPS) of each method. "4K *N/A* on GPU" denotes the compared method is unable to enhance 4K-resolution images on a single NVIDIA RTX 3090 GPU with 24Gb memory. Our method achieves the best PSNR results with faster inference speed (higher FPS) and less computational costs (FLOPs) than the competitors.

more spatial resolution and brightness information. Therefore, it's necessary to utilize the advantages of green channels.

In this paper, we propose ERIENet, an efficient RAW image enhancement network designed to achieve high-quality results with fewer parameters. ERIENet incorporates a multi-scale parallel architecture and Channel-Aware Residual Dense Blocks (CRDB) with channel attention for effective feature extraction. Additionally, we introduce a green channel guidance mechanism that leverages lighting information from the green channel of RGGB RAW images, using spatial adaptive normalization to optimize batch normalization parameters. Tested on SID and ELD dataset, ERIENet demonstrates superior performance in PSNR, SSIM, and visual quality while achieving real-time efficiency with lower memory and computational costs, processing 4K images at 146.2 FPS on the SID dataset. These results highlight its ability to combine high speed and accuracy by optimizing network architecture and utilizing the unique characteristics of RAW data.

## II. Proposed Method

### A. Motivation

Low-light image enhancement transforms dimly lit images into clear, well-lit ones. However, existing methods struggle to balance efficiency and accuracy, especially for high-resolution RAW images, due to sequential processing (*e.g.*,

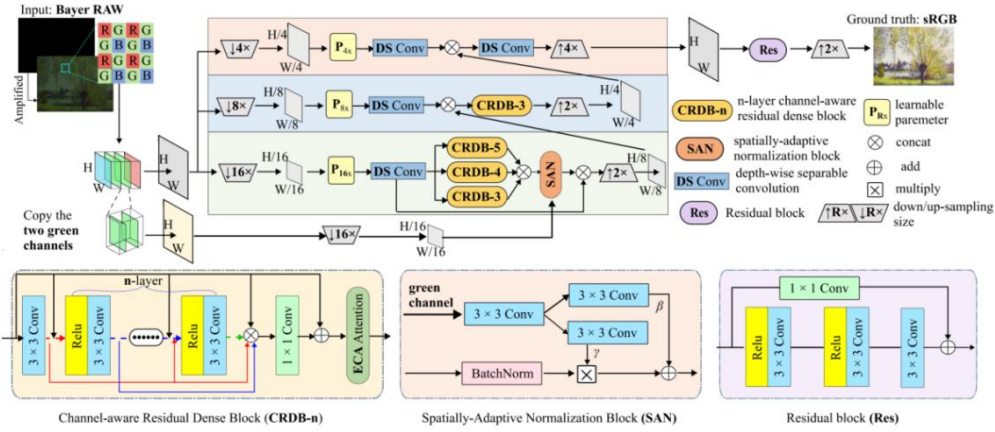

Fig. 2. Overall architecture of our network. The network downsamples the input Bayer pattern (RGGB) RAW image into different scales for pyramid feature learning. We introduce Channel-aware Residual Dense Block (CRDB) of varying depths in the branches at different levels to efficiently extract features at multiple scales. Besides, we propose a Green Channel Guidance (GCG) branch to extract the illumination-sensitive features. Ultimately, we obtain the enhanced sRGB image $I_{out}$ of size $H \times W \times 3$ by stacking a Residual (Res) block.

U-net) and underutilization of RAW data, particularly the green channel in Bayer RAW images. These challenges are further amplified by the demand for lightweight solutions on edge devices. To address these issues, we propose a multi-scale parallel architecture for independent computation across branches, reducing computational time. We introduce channel-aware dense blocks for efficient channel attention and a green channel guidance mechanism to leverage the higher spatial resolution and intensity of green pixels in Bayer RAW images. Together, these innovations enable real-time processing of 4K RAW images with high-quality reconstruction, offering an effective solution for resource-constrained scenarios.

### B. Network Overview

The overall architecture of our network is illustrated in Fig. 2. Our approach consists of three main components. Firstly, a multi-scale parallel feature extraction scheme is employed to extract multi-scale features and implement feature fusion on the multi-scale features. Then, the residual dense block equipped with a channel attention mechanism is utilized to extract the multi-scale features. Moreover, we propose a green channel guidance scheme in this regard to enhance and regulate feature maps of different scales in relation to the primary illumination information captured by the lowest scale. Finally, the enhanced sRGB image is generated by a high-resolution image reconstruction module.

### C. Multi-Scale Parallel Feature Extraction and Fusion

As shown in Fig. 2, our approach begins with a RAW image $I$ ( $H \times W \times 1$ ), transformed into a four-channel (RGGB) image $I_{input}$ ( $H/2 \times W/2 \times 4$ ). This image is processed through $k = 3$ feature extraction branches, where each branch downsamples the input by factors of 4, 8, and 16, respectively, using depth-wise separable convolutions (DS Conv) to reduce computational cost while retaining multi-scale information. For the $i$-th branch, the input $I_{input}$ is downsampled to $I'_{input}$, adjusted by a learnable parameter acting as a globally consistent mask.

Each branch processes features through parallel Channel-aware Residual Dense Blocks (CRDB) to enhance representation and efficiency. Features across scales are progressively fused via upsampling, concatenation, and convolution. Finally, an enhanced sRGB image ($H \times W \times 3$) is generated using a residual block with a skip connection and a × 2 upscaling step.

As depicted in Fig. 3, the multi-scale parallel architecture effectively aggregates pyramid contextual information, producing images with finer textures and fewer artifacts compared to single- or two-scale counterparts. Its key advantages include concentrating computations on the bottom branch for efficiency, leveraging complementary features across scales, and simplifying model learning through progressive fusion.

### D. Channel-aware Residual Dense Block

Residual-dense structures like ResNet, DenseNet, and RDB have good performance in feature extraction but incur high computational costs for high-resolution images. To address this, we propose the Channel-aware Residual Dense Block (CRDB), which integrates Efficient Channel Attention (ECA) into RDBs. A CRDB-$n$, with n ReLU-Conv $3 \times 3$ layers, concatenates outputs from each layer, fuses them via a $1 \times 1$ convolution, and applies ECA with kernel size of 3 for enhanced feature recalibration. This design retains RDB's strengths while reducing computational overhead. In ERIENet, three parallel CRDBs at the third scale replace sequential stacking, improving efficiency without sacrificing performance. CRDB achieves a balance of power and efficiency, making ERIENet practical for real-world low-light enhancement.

### E. Green Channel Guidance

In RAW images, the green channel has twice the sampling rate of red and blue channels, offering higher spatial resolution and stronger intensity sensitivity. To leverage this, we propose a Green Channel Guidance (GCG) branch to extract illumination-sensitive features and guide image reconstruction. Added at the 3rd scale of our network, the GCG branch introduces minimal computational cost. Due to the human eye's sensitivity to green light, camera sensors use Bayer filters with double the green pixels, enhancing brightness perception. Entropy calculations confirm the green channel's superior information content. Exploiting inter-channel dependencies, we use the green channel to recover RGB channels and missing pixel values. The GCG branch employs Spatially-Adaptive Normalization (SAN) to modulate feature extraction in the 3rd scale's CRDBs. Two green channels are processed through $3 \times 3$ convolutions to generate element-wise modulation parameters $\gamma$ and $\beta$, which adjust normalized features for better illumination representation.

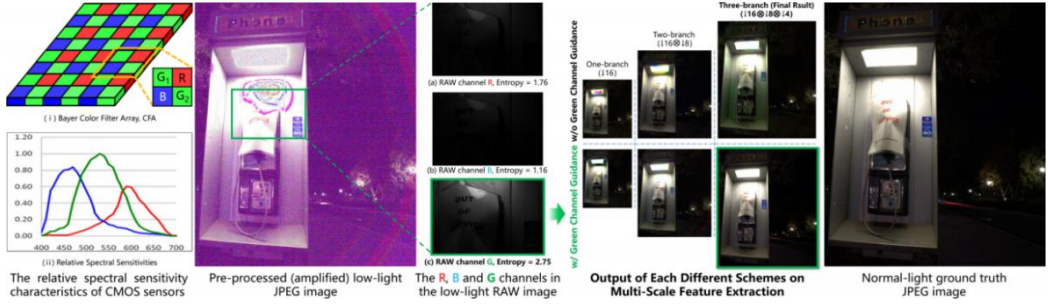

Fig. 3. The study of our network with multi-scale parallel feature extraction architecture and the proposed GCG branch that is based on the relative spectral sensitivity characteristics of CMOS sensors. The output feature map of each scale branch in our network trained with or w/o the proposed GCG branch is provided to validate the effectiveness of our contributions.

The GCG branch improves visual quality by recovering dark area details, reducing noise and color bias, and enhancing texture and color balance. Ablation experiments confirm its effectiveness in low-light image enhancement, demonstrating significant gains in detail and natural color reproduction.

*F. Loss Function*

Traditional pixel-level loss functions (such as $\mathcal{L}_1$ loss) primarily focus on the global error of an image, often failing to effectively constrain the recovery of local high-frequency information. Wavelet transform, through multi-resolution analysis, can simultaneously capture global structure and local details in both the spatial and frequency domains. Inspired by the pioneer wavelet SSIM loss in and the wavelet mean squared error (MSE) loss in, we also utilize the constraint on the wavelet coefficients to emphasize the reconstruction of high-frequency information to enhance the image details.

Specifically, to formulate the wavelet SSIM loss $\mathcal{L}_{wssim}$, we apply two-dimensional Discrete Wavelet Transform (DWT) onto the image $I_{out}$ or $I_{gt}$. Then we can define the wavelet SSIM loss as

$$\mathcal{L}_{wssim} = -\sum_{i=0}^{w,i} r_i SSIM(I_{out}^{w,i}, I_{gt}^{w,i}), \quad (1)$$

where $r_i$ is the $i$-th ratio for multi-frequency SSIM loss, $SSIM(\cdot)$ is the original SSIM loss, and $w \in \{LL, HL, LH, HH\}$. Here, we employ the 2D Haar wavelet and apply the DWT three times in our implementation.

The wavelet MSE loss $\mathcal{L}_{wmse}$ is defined as

$$\mathcal{L}_{wmse} = MSE(I_{out}, I_{gt}) + \sum_{t=1}^{3} MSE(DWT^t(I_{out}), DWT^t(I_{gt})), \quad (2)$$

where $DWT^t(\cdot)$ calculates the $t$-th scale of wavelet coefficients from the $I_{out}$ and $I_{gt}$ images respectively.

In summary, the total loss is the weighted sum of the above three loss functions

$$\mathcal{L} = \mathcal{L}_1 + \lambda_{wssim}\mathcal{L}_{wssim} + \lambda_{wmse}\mathcal{L}_{wmse}. \quad (3)$$

Here, $\lambda > 0$ is used to trade off different terms, and we empirically set $\lambda_{wssim} = \lambda_{wmse} = 0.5$ in all our experiments.

*G. Implement Details*

Our ERIENet is implemented in PyTorch using an Intel Xeon Gold 5218R CPU and NVIDIA RTX 3090 GPU. To handle high-resolution RAW images and avoid memory issues, we pack each RAW image into four channels and crop them into $512 \times 512$ non-overlapping patches for training. Data augmentation includes random flipping and 90°, 180°, and 270° rotations. The model was trained for 500 epochs using the Adam optimizer ($\beta 1=0.5$ and $\beta 2=0.999$) with a batch size of 64 and a base learning rate of $10^{-4}$. For training on the SID datasets, low-light RAW images were used as input, with their corresponding normal-light sRGB images as ground truth. A benchmark was established using Rawpy, a Python wrapper for LibRaw, to process reference RAW images into sRGB ground truth. During inference, full-resolution RAW images were used, with amplification ratios adjusted to match exposure differences between input and reference images, consistent across training and testing.

### III. EXPERIMENT

*A. Dataset and Metric*

**Dataset.** We evaluate our method on the SID [9] and ELD datasets, which provide low-light RAW images with corresponding normal-light RAW images as ground truth. Pre-processed low-light RAW images are used as input to learn color and format conversion without relying on metadata. Ground truth images are generated using a simplified Rawpy pipeline, converting RAW to 8-bit sRGB with proper exposure.

The SID dataset includes images captured with Sony α 7S II ($4256 \times 2848$) and Fujifilm X-Trans sensors, with a focus on the widely used Bayer sensor for consistency. The ELD dataset contains 240 image pairs of 10 indoor scenes captured by cameras like Sony A7S2 and Canon EOS 70D. It includes varying ISO levels (800, 1600, 3200) and exposure settings, replicating diverse noise conditions. This dataset is critical for assessing noise-handling performance in low-light scenarios.

**Metric.** In order to evaluate the performance of compared methods on RAW image low-light enhancement, we use the widely used metrics including Peak Signal-to-Noise Ratio (PSNR), Structural Similarity Index (SSIM, and Learned Perceptual Image Patch Similarity (LPIPS). To compare the complexity of different methods, we evaluate the inference time (including GPU and CPU times), number of floating-point operations per second (FLOPs), and parameter count.

TABLE I. COMPARISON OF DIFFERENT METHODS ON THE INFERENCE TIME (INCLUDING THE GPU AND CPU TIME IN MS), NUMBER OF FLOPs (IN G.), NUMBER OF PARAMETERS (IN M.), PSNR, SSIM, AND LPIPS ON THE SID (SONY BAYER RAW SUBSET) [9] AND ELD DATASET. THE BEST RESULTS ARE HIGHLIGHTED WITH GREEN AND **BOLD**, AND THE SECOND- AND THIRD-BEST RESULTS ARE HIGHLIGHTED IN RED AND BLUE, RESPECTIVELY.

| Dataset | Methods | | Inference Speed / Complexity | | | Performance | | |
|---|---|---|---|---|---|---|---|---|
| | Type | Name | Inference time GPU(ms) | Inference time CPU(s) | GFLOPs | Parameters | PSNR | SSIM | LPIPS |
| SID-Sony (Bayer RAW) | sRGB-based | PJLENR[15] | 525.06 | 8.01 | 113.86 | 2.539 | 22.33 | 0.655 | 0.356 |
| | | DeepLPF[12] | - | 20.80 | 667.54 | 1.769 | 26.62 | 0.771 | 0.264 |
| | | CDAN[16] | 7.46 | 5.53 | 381.44 | 3.586 | 30.05 | 0.811 | 0.263 |
| | | Zero-DCE[14] | - | 9.79 | 967.73 | 0.794 | 25.65 | 0.756 | 0.386 |
| | RAW-based | SID[9] | 9.22 | 5.38 | 523.83 | 7.761 | 28.62 | 0.798 | 0.281 |
| | | REDI[17] | 7.14 | 1.31 | 58.68 | **0.785** | 28.22 | 0.780 | 0.291 |
| | | RAWFormer[10] | 414.94 | 20.69 | 781.54 | 3.401 | 29.22 | 0.790 | 0.258 |
| | | SMG[11] | 494.68 | 7.72 | 1274.27 | 18.355 | 30.17 | **0.834** | **0.238** |
| | | Dnf[13] | - | 47.48 | 2874.44 | 11.140 | **30.62** | 0.797 | 0.343 |
| | | **ERIENet(Ours)** | **6.84** | **1.13** | **39.29** | 1.419 | 29.12 | 0.797 | 0.259 |
| ELD (Bayer RAW) | sRGB-based | PJLENR[15] | 525.06 | 8.01 | 113.86 | 2.539 | 23.25 | 0.569 | 0.364 |
| | | DeepLPF[12] | - | 20.80 | 667.54 | 1.769 | 26.54 | 0.715 | 0.284 |
| | | CDAN[16] | 7.46 | 70.10 | 381.44 | 3.586 | 28.87 | 0.811 | 0.274 |
| | | Zero-DCE[14] | - | 9.79 | 967.73 | 0.794 | 19.84 | 0.268 | 0.602 |
| | RAW-based | SID[9] | 9.22 | 5.38 | 523.83 | 7.761 | 26.44 | 0.783 | 0.291 |
| | | REDI[17] | 7.14 | 1.31 | 58.68 | **0.785** | 26.22 | 0.655 | 0.267 |
| | | RAWFormer[10] | 414.94 | 20.69 | 781.54 | 3.401 | 28.18 | 0.817 | **0.204** |
| | | SMG[11] | 494.68 | 5.53 | 1274.27 | 18.355 | **29.27** | **0.823** | 0.248 |
| | | Dnf[13] | - | 47.48 | 2874.44 | 11.140 | 28.75 | 0.813 | 0.327 |
| | | **ERIENet(Ours)** | **6.84** | **1.13** | **39.29** | 1.419 | 28.80 | 0.819 | 0.281 |

*B. Comparison with State-of-the-arts*

We compare our method with nine state-of-the-art low-light enhancement methods: SID [9], REDI [17], PJLENR [15], SMG [11], Dnf [13], Zero-DCE [14], RAWformer [10], CDAN [16], and DeepLPF [12].

We evaluate ERIENet on the SID and ELD datasets, demonstrating superior performance in both quantitative and qualitative results. On the SID dataset, ERIENet improves PSNR by over 0.5dB compared to most methods, while maintaining low computational costs and fast inference speeds. On the ELD dataset, ERIENet achieves competitive PSNR and SSIM while maintaining a lightweight design and scalability, handling 4K images efficiently. Qualitative results, as shown in Fig. 4, further show that ERIENet produces clearer backgrounds and better detail recovery compared to competitors, with fewer artifacts and more consistent illumination.

*C. Ablation Study*

**Multi-scale parallel feature extraction.** Our method uses a multi-scale parallel framework to extract features by downsampling the input RAW image in Bayer pattern by $4\times$, $8\times$, and $16\times$. The lowest-resolution branch with $16\times$ downsampling is given priority with more convolutional blocks. To evaluate the effectiveness of this framework, we conducted experiments with two variants. One variant uses only the $16\times$ branch, referred to as "↓16". Another variant uses the $16\times$ and $8\times$ branches, referred to as "↓16,8". Our full method, which uses all three branches of $16\times$, $8\times$, and $4\times$ downsampling, is referred to as "↓16,8,4". Results on the SID dataset show that our full method achieves significantly better performance in terms of PSNR, SSIM, and LPIPS. This demonstrates the importance of parallel multi-scale feature extraction for efficient low-light RAW image enhancement.

**Green channel guided normalization.** We utilize the green channel to guide feature extraction in the $16\times$ branch by incorporating the GCG branch with Spatially-Adaptive Normalization (SAN). To evaluate its effectiveness, we design two variants. The first variant, "None w/ BN", removes the GCG branch and replaces SAN with Batch Normalization (BN). The second variant, "GCG w/ LN", replaces BN in SAN with Layer Normalization (LN). On the SID dataset, our method with "GCG w/ BN" outperforms "None w/ BN" by 1.01dB in PSNR, 0.040 in SSIM, and 0.098 in LPIPS, demonstrating the GCG branch's effectiveness for RAW low-light enhancement, and the batch dimension is crucial for capturing useful information in low-light image enhancement.

**Channel-aware Residual Dense Block.** Here, we illustrate the structure and effectiveness of our proposed CRDB. To evaluate its performance, we replace the CRDB-n block with other variants, including the standard Residual Block (RB), Dense Block (DB), Residual Dense Block (RDB), and its improved version RDB*, all with comparable parameters.

As summarized in TABLE IV, CRDB, with its efficient channel attention mechanism, achieves superior performance and efficiency for low-light RAW image enhancement.

IV. CONCLUSION

In this paper, we propose a multi-scale parallel feature extraction network, ERIENet, for efficient and high-performance low-light enhancement of Bayer pattern RAW images. Our fully parallel architecture incorporates an efficient channel-aware residual dense block, enabling compact yet dense feature extraction with lower computational costs, significantly improving inference speed for 4K images. Additionally, we utilize the green channels of RAW images in the lowest-resolution branch to guide overall feature extraction. Experiments on the SID and ELD dataset show that ERIENet outperforms previous state-of-the-art methods in speed and efficiency while achieving strong results in objective metrics and visual quality. Ablation studies confirm that the GCG branch enhances feature extraction and pixel recovery, leading to better low-light enhancement performance. Although ERIENet excels in single-frame enhancement, exploring frame co-relationships in low-light videos remains an open challenge. Furthermore, low-light enhancement could be integrated into tasks like object detection, semantic segmentation, and monocular depth estimation. We hope our findings promote advancements in both high- and low-level visual tasks.

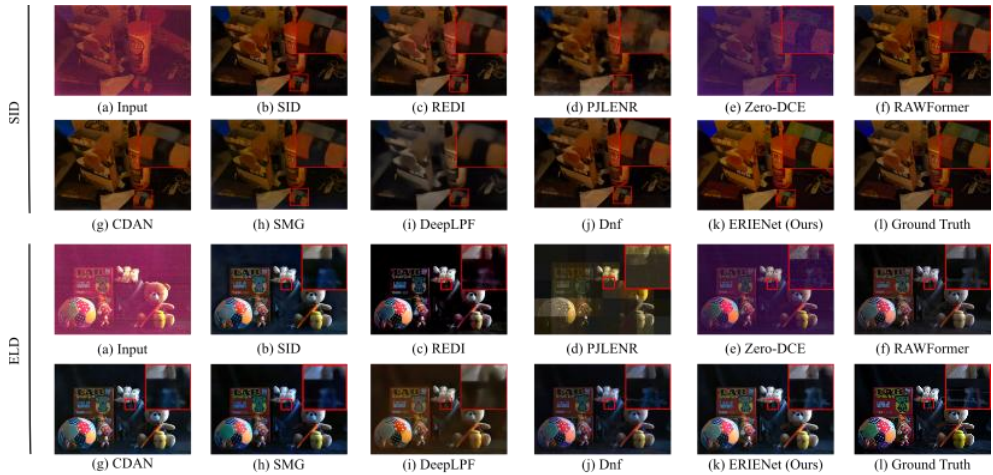

Fig. 4. **The visual quality comparison** of enhanced images by different methods tested on the SID dataset and ELD dataset. The red boxes represent the saliency regions of the results. (Best viewed on screen with zoom).

TABLE II. PERFORMANCE OF OUR METHOD WITH DIFFERENT SCHEMES ON MULTI-SCALE PARALLEL FEATURE EXTRACTION.

| Scale | Time(ms) | GFLOPS | PSNR ↑ | SSIM ↑ | LPIPS ↓ |
|---|---|---|---|---|---|
| ↓16 | 5.25 | 27.75 | 27.86 | 0.757 | 0.363 |
| ↓16,8 | 6.49 | 36.70 | 28.15 | 0.772 | 0.305 |
| ↓16, 8, 4(Ours) | 6.84 | 39.29 | **29.12** | **0.797** | **0.259** |

TABLE III. COMPARISON OF DIFFERENT VARIANT OF OUR METHOD.

| Guidance Scheme | Time(ms) | GFLOPS | PSNR ↑ | SSIM ↑ | LPIPS ↓ |
|---|---|---|---|---|---|
| None w/ BN | 6.62 | 34.63 | 27.97 | 0.758 | 0.365 |
| GCG w/ LN | 6.95 | 39.29 | 28.02 | 0.765 | 0.301 |
| GCG w/ BN (Ours) | 6.84 | 39.29 | **29.12** | **0.797** | **0.259** |

TABLE IV. CRDB REMOVE ECA; REPLACE CRDB BY RB OR DB (BUT WITH COMPARABLE PARAMETER AMOUNT).

| Varient | Time(ms) | GFLOPS | PSNR ↑ | SSIM ↑ | LPIPS ↓ |
|---|---|---|---|---|---|
| RB | 5.65 | 26.86 | 27.94 | 0.768 | 0.301 |
| DB | 6.67 | 39.88 | 28.57 | 0.779 | 0.278 |
| RDB | 6.72 | 39.85 | 28.66 | 0.780 | 0.281 |
| RDB* | 6.82 | 39.88 | 28.79 | 0.787 | 0.277 |
| CRDB (Ours) | 6.84 | 39.29 | **29.12** | **0.797** | **0.259** |

REFERENCES

[1] H. Li and Y. Fu, "FCDFusion: A fast, low color deviation method for fusing visible and infrared image pairs," in Computational Visual Media, vol. 11, no. 1, pp. 195–211, 2025.

[2] H. Li, Z. Wu, R. Shao, T. Zhang, and Y. Fu, "Noise calibration and spatial-frequency interactive network for STEM image enhancement," in CVPR, pp. 21287–21296, June 2025.

[3] Y. Zhang, Z. Lai, T. Zhang, Y. Fu, and C. Zhou, "Unaligned RGB guided hyperspectral image super-resolution with spatial-spectral concordance," in International Journal of Computer Vision, vol. 133, no. 9, pp. 6590–6610, 2025.

[4] T. Zhang, Y. Fu and J. Zhang, "Deep Guided Attention Network for Joint Denoising and Demosaicing in Real Image," in Chinese Journal of Electronics, vol. 33, no. 1, pp. 303-312, January 2024

[5] Y. Li et al., "Lightweight Object Detection Networks for UAV Aerial Images Based on YOLO," in Chinese Journal of Electronics, vol. 33, no. 4, pp. 997-1009, 2024.

[6] H. YANG, H. LIU, Y. ZHANG, et al., "FMR-GNet: Forward Mix-Hop Spatial-Temporal Residual Graph Network for 3D Pose Estimation," in Chinese Journal of Electronics, vol. 33, no. 6, pp. 1346–1359, 2024

[7] J. Zheng, B. Jiang, W. Peng and Q. Zhang, "Multi-Scale Binocular Stereo Matching Based on Semantic Association," in Chinese Journal of Electronics, vol. 33, no. 4, pp. 1010-1022, July 2024

[8] L. Chen, Y. Fu, L. Gu, D. Zheng and J. Dai, "Spatial Frequency Modulation for Semantic Segmentation," in IEEE Transactions on Pattern Analysis and Machine Intelligence, vol. 47, no. 11, pp. 9767-9784, Nov. 2025

[9] C. Chen, Q. Chen, J. Xu, and V. Koltun, "Learning to see in the dark," in CVPR, 2018, pp. 3291–3300.

[10] X. Wei, X. Dong, L. Ma, B.-J. Teoh, and Z. Lin, "Rawformer: An efficient transformer-based model for raw image restoration," in IEEE Signal Processing Letters, vol. 29, pp. 2677–2681, 2022.

[11] R. Xiaojiang and J. Luo, "Low-light image enhancement via robust retinex model and noise," in CVPR, 2020.

[12] E. Moran, P. Marza, S. McDonagh, S. Parisot, and G. Slabaugh, "Noir: Noise insensitive unsupervised raw image reconstruction," in CVPR, 2020, pp. 12826–12835.

[13] K. Xu, H. Chen, C. Xu, Y. Jin, and C. Zhu, "Decouple and enhance: Towards high-quality low-light image enhancement," in IEEE Transactions on Circuits and Systems for Video Technology, 2022.

[14] C. Guo, C. Li, J. Guo, C. C. Loy, J. Hou, S. Kwong, and R. Cong, "Zero-reference deep curve estimation for low-light image enhancement," in IEEE Transactions on Image Processing, vol. 29, pp. 6284–6295, 2020.

[15] M. Young, "The Technical Writer's Handbook." in Mill Valley, CA: University Science, 1989.

[16] H. Shakibania, S. Raoufi, and H. Khotanlou, "CDAN: Convolutional dense attention-guided network for low-light image enhancement," in Digital Signal Processing, vol. 156, p. 104802, 2025.

[17] M. Lamba and P. Mitra, "Restoring extremely dark images in real time," in CVPR, 2021, pp. 9924–9933.